RESEARCH ARTICLE

# A semi-automated segmentation method for detection of pulmonary embolism in True-FISP MRI sequences

Soenksen LR[a], Jiménez-Angeles L[b], Melendez G[c], Meave A[c]

[a] Biomedical Engineering Department, Instituto Tecnológico de Estudios Superiores de Monterrey Campus Ciuidad de México, Mexico City, México.
[b] Nuclear Cardiology Department, Instituto Nacional de Cardiología Ignacio Chávez, Mexico City, México.
[c] MRI Department, Instituto Nacional de Cardiología Ignacio Chávez, Mexico City, México.

Pulmonary embolism (PE) is a highly mortal disease, currently assessed by pulmonary CT angiography. True-FISP MRI has emerged as an innocuous alternative that does not hold many of the limitations of x-ray imaging. However, True-FISP MRI is very sensitive to turbulent blood flow, generating artifacts that may resemble fake clots in the pulmonary vasculature. These misinterpretations reduce its overall diagnostic accuracy to 94%, limiting a wider use in clinical environments. A new segmentation algorithm is proposed to confirm the presence of real pulmonary clots in True-FISP MR images by quantitative means, measuring the shape, intensity and solidity of the formation. The algorithm was evaluated in 37 patients. The developed method increased the diagnostic accuracy of expert observers assessing Pulmonary True-FISP MRI sequences by 6% without the use of ionizing radiation, achieving a diagnostic accuracy comparable to standard CT angiography.

## 1. INTRODUCTION

Pulmonary embolism (PE) is a severe cardiovascular condition defined by the formation of thrombus in the pulmonary veins, which affects 6 to 7 individuals per 10000 every year globally.[1] It is also a condition that holds a significant mortality rate in the developed world, especially in North America (3.5% to 15%).[2,3] where almost 65% of the patients with candid suspicion of PE will die within the first hour of symptomatic appearance, and 92.9% of the rest will likely do so in less than 2.5 hours.[4] Rapid diagnostics and confirmatory tests for PE are critical to reduce the burden of this condition, and significant research has been done to improve its diagnosis and treatment.

Diverse imaging modalities have been proposed to evaluate PE, including ultrasound, ventilation / perfusion (V/Q) scintigraphy, computerized tomography (CT) and magnetic resonance imaging (MRI). Unfortunately, each modality presents significant challenges of availability, and accuracy. The Prospective Investigation of Pulmonary Embolism Diagnosis study[5] concluded that (V/Q) scintigraphy is not conclusive enough to provide an accurate diagnosis, and the TUSPE study showed that negative chest ultrasound result does not rule out a PE, therefore requiring a confirmatory test.[6,7] These problems permeate even the gold standard for evaluating PE: CT pulmonary angiography (CTPA), where up to 25% of patients that could be diagnosed by this technique present one or more contraindications for x-ray scanning and cannot undergo this imaging procedure.[8,9,10,11,12] Some of the key contraindications presented by many PE patients include pregnancy, renal failure and allergic reactions to iodide contrast agents.[5,13]

Recently, specialized MRI sequences have been introduced as an appealing option to screen and diagnose previously contraindicated patients with PE, with high-resolution outputs.[14] One of such sequences is called: True fast Imaging with Steady State Precession (True-FISP) MRI, which can be used to visualize vascular structures by non-ionizing means and without the use of any contrast medium.[15] Even though considerable research has been done in this field, it is still believed that True-FISP MRI holds an accuracy value of approximately 93%-97% or 5% less than CT (gold standard) due to diagnostic misinterpretations.[16] Both, real clot formations and low-density regions of turbulent blood flow in True-FISP MRI are prone to exhibit similar hypointense patterns within segmental and sub-segmental arteries, problem that has limited its use in suspicion of PE. It is the aim of the present work to introduce a semi-automated segmentation method capable of reducing the -diagnostic uncertainties- related to clot detection in pulmonary True-FISP MRI. Experimental results on the implementation of said method are presented to evaluate its clinical application.

## 2. MATERIALS AND METHODS

All image transformations and calculations were done using MATLAB® (The MathWorks, Inc.; Natick, Massachusetts) and the MATLAB® image processing toolbox. Through visual evaluations in True-FISP images we noticed that spurious clot-like formations in non-trombolized arteries exhibited anomalous parameters in terms of solidity, eccentricity and mean intensity value. These parameters were proposed by the investigation team as a way to distinguish real clot images from fictitious formations caused by turbulent flow in the pulmonary vasculature. The proposed method consists in a series of filtering processes to prepare the image for quantitative analysis. Then, a gray-scale threshold based segmentation is applied, followed by an intermediate morphological analysis and a conditional dichotomist characterization based on the previously mentioned parameters.

A descriptive chart regarding the proposed algorithm is showed in figure 2. The filtering procedures and the morphological transformations used for preparing the quantitative analysis are explained in the filtering and transformation section. The anatomical characterization process (ROI placement), sequential usage of filtered images and decision-making process are explained in the general algorithm section.

### 2.1. *General process*

#### 2.1.1. *Filtering and transformation*

The analysis of low intensity regions of luminal zones in True-FISP MRI (possible clots) required the application a series of linear filters in order to eliminate diffuse segments and to enhance significant structures within the vessel. First, an unsharp contrast enhancement filter is applied to obtain an image with enhanced border information.[17] The function for this filter is presented in equation 1. The constant λ defines the effect of the filter and ranges between 0 and 1. For this application, the value of λ was set to 0.21 to obtain a moderate degree of sharpening.

$$f_{sharp}(m,n) = f(m,n) + \lambda * g(m,n) \qquad (1)$$

The sharpened image is then transformed using a Contrast-limited Adaptive Histogram Equalization (CLAHE).[18] The CLAHE transformation is specified to operate on small regions of the image (tiles), enhancing each tile individually. The neighboring tiles were combined using a bilinear interpolation to eliminate artificially induced boundaries. This resultant image is then superposed to the original DICOM image in two different ways. a) A linear addition of both images, and b) A weighted additive transformation of those images (with a 2 to 1 relationship). This filtering image process is shown in the Figure 1. The last two filtered images keep low frequency components while enhancing most of the edges. It is important to mention that the resultant images must be re-equalized, by a second CLAHE transformation, to obtain better contrast range.

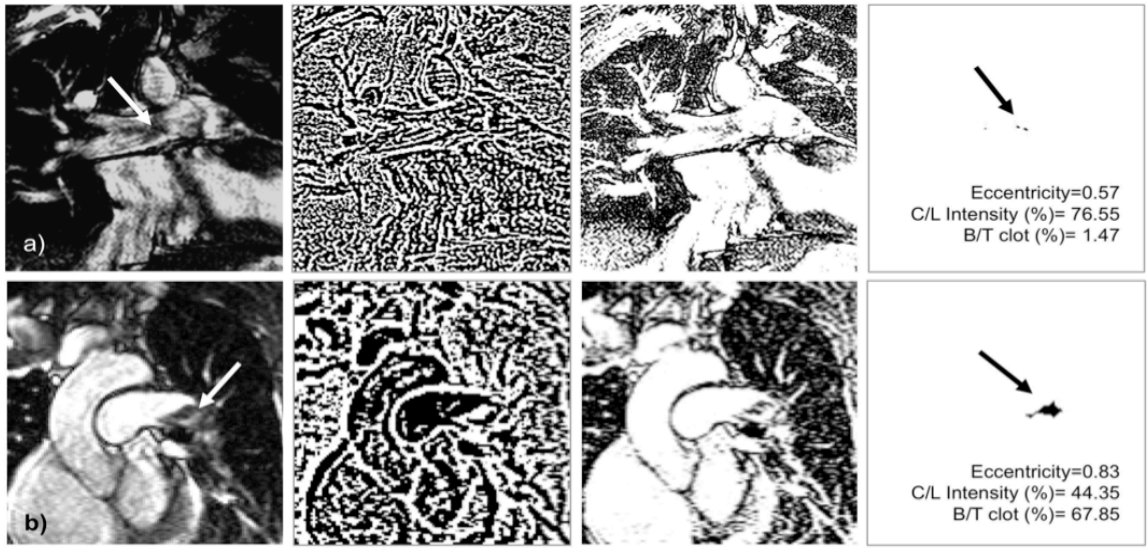

**Figure 1**.- Filtering and quantification process, in a) a clot-like formation is located (arrow), the image is filtered and correctly characterized as NEGATIVE for PE by the algorithm. In b) the located clot-like formation (arrow) is quantified and characterized as POSITIVE for PE after undergoing the filtering process.

*2.1.2. Clot Characterization*

Once the mentioned filters have been applied, a characterization step operates to determine whether the selected hypo-intense region is a real clot or not. In the initial stage of the algorithm workflow (Figure 2), the user must draw two different regions of interest (ROIs). One ROI should correspond to the luminal section of a main pulmonary artery and the other should delimitate the clot-like body. Those selected regions are compared to corroborate that the clot region is completely located within the vessel lumen. If the previous condition is true, three binary images are created in correspondence to: 1) the clot-like structure, 2) the arterial lumen and 3) the arterial lumen without the clot selected region.
From these binary masks, the algorithm selects and merges the clot-segmented region with the simple enhanced filtered image, and does the same for the luminal region. After an additive processes, the algorithm calculates the mean intensity ratio of the clot-like structure compared to the mean intensity in luminal-only zone. A mean intensity range for real-clot formations of 40±20% intensity units (IU) compared to the mean lumen intensity is proposed, to provide the best results in the present method. The measurement of occupational rate is done using the previously calculated binary masks and the last weighted filtered image. Within a merged version of both images a gray-scale threshold is applied to binarize it according to the intensities of the inner clot region. The Otsu's method is used for this purpose[19], automatically choosing a specific threshold value to minimize the variance of the segmented image. As weighted images in this application acquire depleted low frequency components and border enhancement, bogus clot structures tend to be eliminated and therefore no binary structure appears inside the clot-like region.

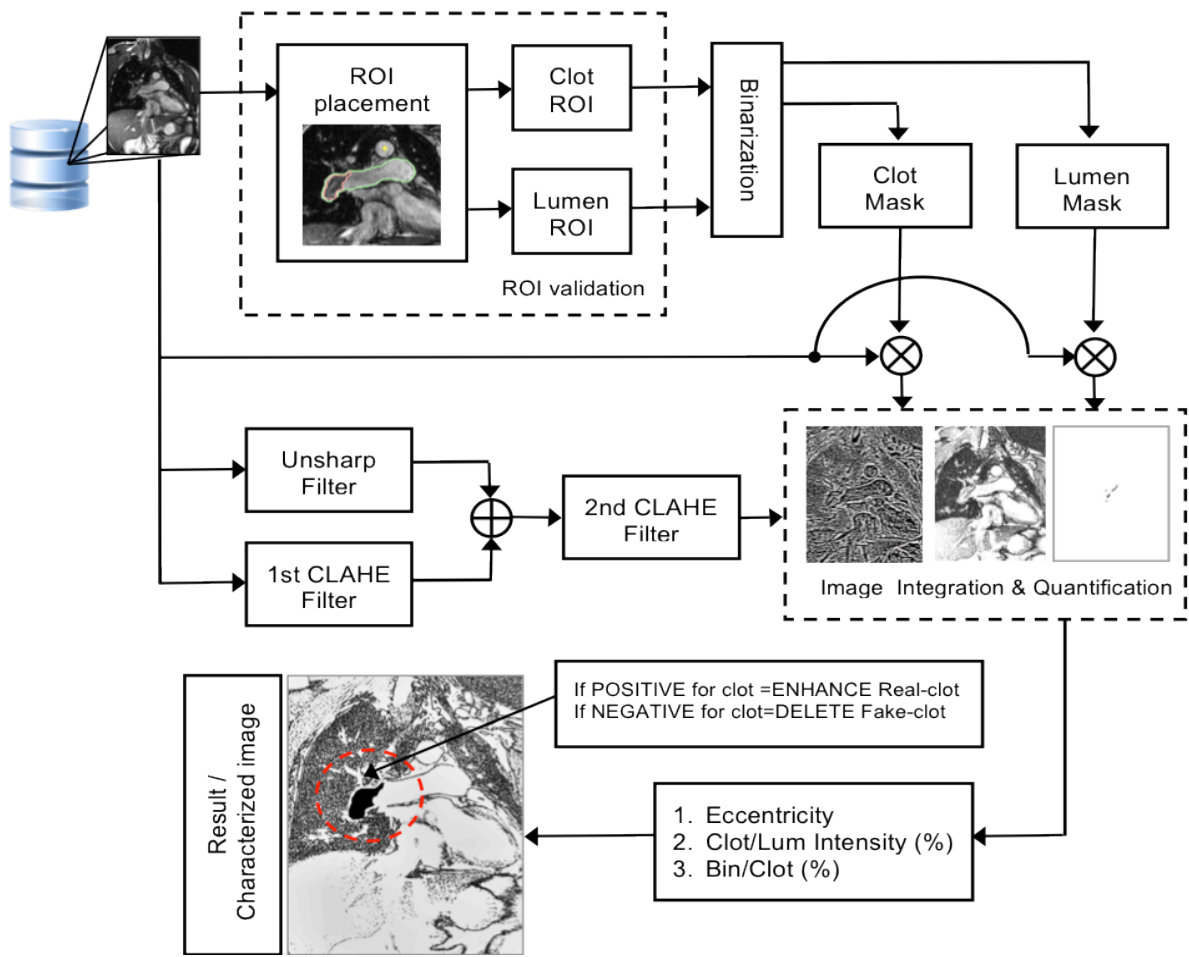

**Figure 2**.- Algorithm block diagram. After DICOM image recuperation, and ROI placement, binarization and filtering is done in a parallel way. The results of binarization and filtering are used to generate a series of enhanced or segmented images for posterior quantification and characterization.

If one or more solid binary regions appear in the image, a morphological closing operation is done with a 5-pixel disk-structuring element. The algorithm calculates the area of all solid structures and compares them to the entire clot-like area selected by the user. If the solid structure area exceeds 7% of the total selected clot-like region, this parameter indicates high suspicion of PE. This parameter was proposed through empirical evaluation and showed to include small clot formations. The eccentricity is the third evaluated parameter and is a region property that describes the shape of a structure in terms of likeliness to be a circle or a line. The eccentricity parameter is calculated from the binary image of the clot-segmented region. For our application, possible eccentricity values can range from 0 to 1, being "0" a perfect circle and "1" a line. The region eccentricity proposed to be indicative of real clot if <0.8. This threshold was selected empirically under the assumption that real clot formations are the result of concentrically aggregation of small thrombi that will result in a more circular shape than the shape usually adopted by fluid turbulence.

All the threshold values proposed above were adjusted using independent True-FISP MRI studies (3) not used in later validations. Once the parameters have been evaluated, the algorithm uses this information to indicate the presence or absence of a real clot. Real clot structures were observed to have positive values in two or more of these parameters. If mean intensity and occupational values were indicative of a real clot, the results were established as positive regardless eccentricity. If only one parameter was indicative of PE, circular eccentricity was required conclude this diagnosis.

## 3. ANALYSIS AND EVALUATION

### 3.1. *Protocol and Patients*

In order to show the effectiveness of the proposed algorithm a series of clinical evaluations were carried out. Thirty-seven patients with different pretest probability scores for PE (according to Wells and Geneva criteria) were included in the study.[20,21,22] All the patients were > 18 years old and had clinically suspicion for acute PE.[23] Of the complete group of patients 15 were men (40.5%) and 22 were women (59.5%), with an age range of 48.4 ± 16.3 years. All CT and MR studies were obtained consecutively during periods of staff availability, daytime weekdays. The prevalence of the disease in the group was 16.2%. Pulmonary CT angiography and True-FISP MRI were done with at most 6 hours of difference in order to obtain analogous studies in both formats.

### 3.2. *Visual CT pulmonary angiography and MRI True-FISP analysis*

Pulmonary CT angiographies were acquired with a 64-detector CT scanner (Siemens Somatom Sensation 64, Forcheim, Germany). The acquisition was made with a collimation of 24 × 1.2 mm, gantry rotation time of 330 ms @400 mAs /100 kV with cardiac modulation and bolus tracking acquisition. The CT images were reconstructed using a kernel-b30f filter and were visually evaluated by two experts in the area: GM (6 years of experience) and AM (8 years of experience). All MRI True-FISP images were obtained with a 1.5 T MRI system (Siemens Sonata, Forcheim, Germany). The acquisition was done with a 12-slice scheme with a FOV 400 mm, TR: 295.81 ms, TE: 1.16 ms, Echo spacing=2.7 ms using a body- spin antenna and a flip angle of 70º. Two images were selected from each MRI study, one that showed the entire left pulmonary artery and one of the right pulmonary artery. The slice matrix was established to be 256 x 256 mm2, with slice thickness of 4.0 mm and voxel size of 1.8mm x 1.8mm x 4.0mm. By CT pulmonary angiography (gold standard), it was determined that 6 patients (16.2%) were positive for PE in one or more segmental or subsegmental arteries. In the case of the True-FISP MRI analysis, the sequences were visually assessed by three minimally trained evaluators (LS, FM, GL) with no previous reference of the clinical outcome of the patients showing Accuracy=91.8%, Sensitivity=83.3%, Specificity=93.5%, PPV=71.4% and NPV=96.7%. These results indicated the presence of 2 false positive and 1 that was false negative compared to the standard.

## 4. RESULTS AND DISCUSSION

After using the proposed tool all True-FISP MRI studies were characterized correctly by the same minimally trained evaluation group, suggesting that the method was indeed useful to reduce misdiagnosis of PE in True-FISP MRI studies, while being relatively independent of the expertise of the evaluator. Previously true-negatives (Figure 3) were still negative; previous true-positive studies were also segmented as positive (Figure 4), and previously false-positive studies were correctly evaluated as negative (Figure 5).

Even though, the proposed method showed to effectively increase the diagnostic accuracy of pulmonary True-FISP MRI in PE diagnosis to make it comparable to CT, there were some limitations that could have affected these results. The segmentation results are inherently affected by the evaluator's placement of the luminal and clot-like ROIs, as they are not automatically obtained. It was also observed that in order to avoid miscalculation in shape, occupational ratio and mean intensity, the clot-like ROI must be as elliptical as possible. In terms of the validation group, the number of available false-positive and false-negative True-FISP MRI studies was reduced (only 8.1% of the total validation sample). Moreover, the sample size and distribution of said group might not be ideal, since only 37 studies were available from which only 16.2% were PE positive. Still, it is important to mention that this is the first time such a method has been proposed to enhance the diagnostic accuracy of True-FISP MRI in suspicion of PE; as well as the introduction of three novel parameter thresholds to be used in clot characterization (values in solidity, eccentricity and mean intensity in standard MRI DICOM images). Finally, we encourage further investigation of the proposed method, parameters and thresholds to better validate the present results and to accurately define its diagnostic value.

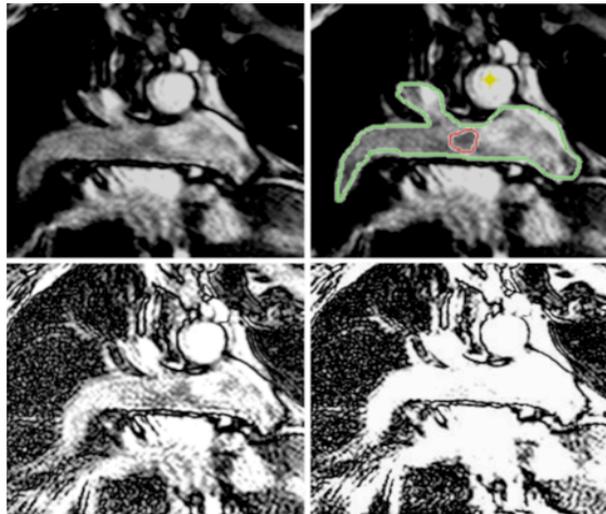

**Figure 3**.- Study case of right pulmonary artery with diffuse low intensity zones within the lumen. This case was assessed as PE negative by the visual and quantitative methods.

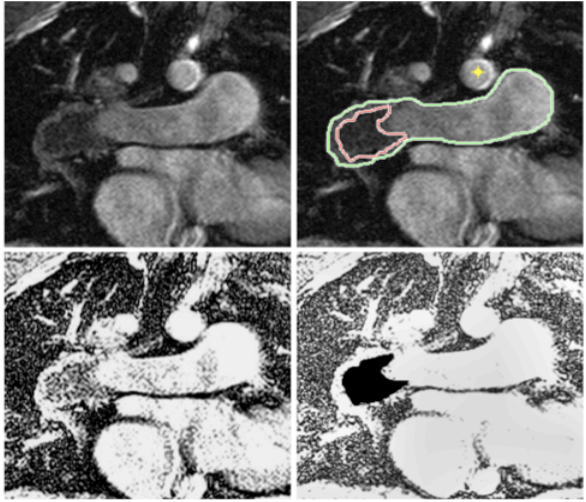 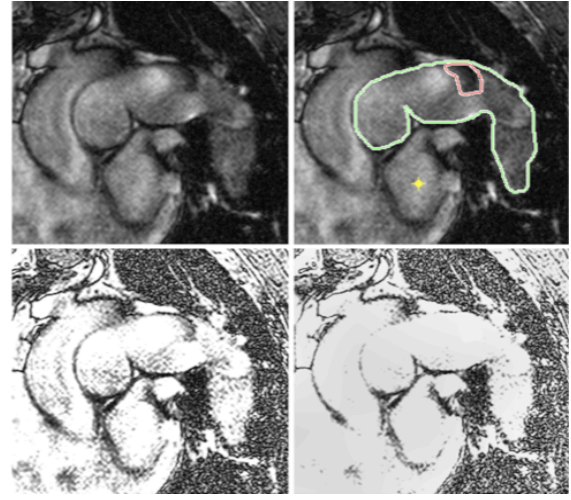

**Figure 4**.- Study case of right pulmonary artery with a well marked low intensity zone within the lumen. This case was assessed as PE positive by the visual and quantitative methods.

**Figure 5**.- Study case of right pulmonary artery with a significant low intensity zone within the lumen. This case was assessed as PE positive by the visual and correctly assessed as negative by the proposed method.

## V. REFERENCES

[1] Warren, D. J., & Matthews, S. (2012). Pulmonary embolism: investigation of the clinically assessed intermediate risk subgroup. *The British journal of radiology, 85(1009)*, 37-43. doi:10.1259/bjr/17451818
[2] Stein PD, Hull RD, Ghali WA, et al. Tracking the uptake of evidence: two decades of hospital practice trends for diagnosing deep vein thrombosis and pulmonary embolism. *Arch Intern Med*. 2003;163:1213-1219.
[3] Horlander KT, Mannino DM, Leeper KV. Pulmonary embolism mortality in the United States, 1979-1998: an analysis using multiple-cause mortality data. *Arch Intern Med*. 2003;163:1711-1717.
[4] Stein PD, Henry JW. Prevalence of acute pulmonary embolism among patients in a general hospital and at autopsy. *Chest* 1995;108:978–981.
[5] PIOPED-Investigators. Value of ventilation/perfusion scan in acute pulmonary embolism. JAMA 1990; 263:2753–2759
[6] Mathis, G., Blank, W., Reissig, A., Lechleitner, P., Reuss, J., Schuler, A., & Beckh, S. (2005). Thoracic ultrasound for diagnosing pulmonary embolism: a prospective multicenter study of 352 patients. *Chest, 128(3)*, 1531-8. doi:10.1378/chest.128.3.1531
[7] Mathis, G. (2009). Two silver standards in the imaging of pulmonary embolism. *Ultraschall in der Medizin (Stuttgart, Germany : 1980), 30(5)*, 497-8; author reply 498-9. doi:10.1055/s-0028-1109926
[8] Schoepf U, Goldhaber S and Costello P. Spiral Computed Tomography for Acute Pulmonary Embolism. *Circulation* 2004;109:2160-2167
[9] Stein PD, Fowler SE, Goodman LR and PIOPED II Investigators. Multidetector computed tomography for acute pulmonary embolism. *N Engl J Med*. 2006 Jun 1;354(22):2317-27.
[10] Wittram C, Maher M, Yoo A, et al. CT Angiography of Pulmonary Embolism: Diagnostic Criteria and Causes of Misdiagnosis. *RadioGraphics* 2004;24:1219–1238.
[11] Kanne J, Lalani T. Role of Computed Tomography and Magnetic Resonance Imaging for Deep Venous Thrombosis and Pulmonary Embolism. *Circulation*. 2004;109[suppl I]:I-15-I-21.
[12] Riedel M. Venous thromboembolic disease. Acute pulmonary embolism 1: Pathophysiology, clinical presentation, and diagnosis. *Heart* 2001;85:229–240.
[13] Clemens S, Leeper K Jr. Newer Modalities for Detection of Pulmonary Emboli. *The American Journal of Medicine* (2007) Vol 120 (10B), S2–S12.
[14] Oudkerk, M., van Beek, E. J. R., Wielopolski, P., van Ooijen, P. M. A., Brouwers-Kuyper, E. M. J., Bongaerts, A. H. H., & Berghout, A. (2002). Comparison of contrast-enhanced magnetic resonance angiography and conventional pulmonary angiography for the diagnosis of pulmonary embolism: a prospective study. *Lancet,*

*359(9318),* 1643-7. doi:10.1016/S0140-6736(02)08596-3

[15] Fuchs, F., Laub, G., & Othomo, K. (2003). TrueFISP—technical considerations and cardiovascular applications. *European Journal of Radiology, 46(1),* 28-32. doi:10.1016/S0720-048X(02)00330-3

[16] Kluge, A., Müller, C., Hansel, J., Gerriets, T., & Bachmann, G. (2004). Real-time MR with TrueFISP for the detection of acute pulmonary embolism: initial clinical experience. *European radiology, 14(4),* 709-18. doi:10.1007/s00330-003-2164-5

[17] Yan Pore Y, Ibrahim H, Kong N; "Investigation on several methods to correct the intensity inhomogeneity in magnetic resonance images," *Innovative Technologies in Intelligent Systems and Industrial Applications, 2008. CITISIA 2008. IEEE Conference on* , vol., no., pp.55-59.

[18] Yang Yu, Hong Zhao; , "A Texture-based Morphologic Enhancement Filter in Two-dimensional Thoracic CT scans," Networking, Sensing and Control, 2006. ICNSC '06. Proceedings of the 2006 IEEE International Conference on , vol., no., pp.850-855, 0-0 0.

[19] Minghui Z, Zhentai L; "Image segmentation based on mutual information," *Medical Image Analysis and Clinical Applications (MIACA), 2010 International Conference on* , vol., no., pp.71-74

[20] Iles S, Hodges A, Darley J, et al. Clinical experience and pre-test probability scores in the diagnosis of pulmonary embolism. *QJMed* 2003; 96:211–215

[21] Wells P, Anderson D, Rodger M, et al. Derivation of a simple clinical model to categorize patients' probability of pulmonary embolism: increasing the models utility with the SimpliRED D-dimer. *Thromb Haemost.* 2000; 83: 416–420.

[22] Wicki J, Perneger T, Junod A, et al. Assessing clinical probability of pulmonary embolism in the emergency ward: a simple score. *Arch Intern Med.* 2001; 161:92–97.

[23] Jerjes-Sánchez C, Elizalde J, Sandoval J, et al. Diagnóstico, estratificación y tratamiento de la tromboembolia pulmonar aguda. Guías y Recomendaciones del Capítulo de Circulación Pulmonar de la Sociedad Mexicana de Cardiología. *Archivos de Cardiología de México;* 2004; 74:S547-S585.